\documentclass[letterpaper, 10 pt, conference]{ieeeconf}
\usepackage{cite}
\usepackage{amsmath,amssymb,amsfonts}
\usepackage{algorithmic}
\usepackage{graphicx}
\usepackage{textcomp}
\usepackage{tabularx}  % for 'tabularx' environment and 'X' column type
\usepackage{ragged2e}  % for '\RaggedRight' macro (allows hyphenation)
\usepackage{makecell}  % for '\makecell' macro
\usepackage{array} % Add this line to your preamble
\usepackage{mdframed}
\usepackage{comment}
\usepackage{url}
\usepackage[T1]{fontenc}
\usepackage{csquotes}
\usepackage{float}
\usepackage{tabularx}  % For 'tabularx' environment
\usepackage{booktabs}  % For \toprule, \midrule, \bottomrule
\usepackage{multirow}  % For \multirow command
\usepackage{hyperref}

\newcolumntype{L}{>{\RaggedRight\arraybackslash}X} % modified 'X' column type

\def\BibTeX{{\rm B\kern-.05em{\sc i\kern-.025em b}\kern-.08em
    T\kern-.1667em\lower.7ex\hbox{E}\kern-.125emX}}

\IEEEoverridecommandlockouts
\title{\LARGE \bf
GPT-4V(ision) for Robotics: Multimodal Task Planning from Human Demonstration
}
\author{
Naoki Wake$^{1}$,
Atsushi Kanehira$^{1}$,
Kazuhiro Sasabuchi$^{1}$,
Jun Takamatsu$^{1}$,
and Katsushi Ikeuchi$^{1}$
\thanks{$^{1}$Applied Robotics Research, Microsoft, 
        Redmond, WA 98052, USA
        {\tt\small naoki.wake@microsoft.com}}%
}

\begin{document}
\maketitle
\thispagestyle{empty}
\pagestyle{empty}
\begin{abstract}

We introduce a pipeline that enhances a general-purpose Vision Language Model, GPT-4V(ision), to facilitate one-shot visual teaching for robotic manipulation. This system analyzes videos of humans performing tasks and outputs executable robot programs that incorporate insights into affordances. The process begins with GPT-4V analyzing the videos to obtain textual explanations of environmental and action details. A GPT-4-based task planner then encodes these details into a symbolic task plan. Subsequently, vision systems spatially and temporally ground the task plan in the videos---objects are identified using an open-vocabulary object detector, and hand-object interactions are analyzed to pinpoint moments of grasping and releasing. This spatiotemporal grounding allows for the gathering of affordance information (e.g., grasp types, waypoints, and body postures) critical for robot execution. Experiments across various scenarios demonstrate the method's efficacy in enabling real robots to operate from one-shot human demonstrations. Meanwhile, quantitative tests have revealed instances of hallucination in GPT-4V, highlighting the importance of incorporating human supervision within the pipeline. The prompts of GPT-4V/GPT-4 are available at this project page: \href{https://microsoft.github.io/GPT4Vision-Robot-Manipulation-Prompts/}{https://microsoft.github.io/GPT4Vision-Robot-Manipulation-Prompts/}
\end{abstract}

\section{Introduction}
\label{sec:introduction}
In light of the substantial progress in Large Language Models (LLM) and Vision Language Models (VLM) in recent years, a number of methodologies have emerged that convert language/visual inputs into robotic manipulation actions. While a mainstream approach is training custom models based on extensive data of robot actions~\cite{jiang2022vima,brohan2023rt,brohan2022rt,li2023vision,ahn2022can,shah2023mutex,li2023mastering}, several studies have explored the use of general-purpose, off-the-shelf language models such as ChatGPT~\cite{OpenAI} and GPT-4~\cite{OpenAIgpt4} through prompt engineering without additional training~\cite{wake_chatgpt,huang2022language,xu2023creative,zhou2023generalizable,ni2023grid,li2023interactive,hu2023tree, vemprala2023chatgpt}. One key advantage of using off-the-shelf models is their flexibility; they can be adapted to various robotic hardware configurations and functionalities simply by modifying prompts. This approach removes the necessity for extensive data collection and model retraining for different hardware or scenarios, greatly improving system reusability in research and easing the transition to industrial applications. Hence, utilizing off-the-shelf models for robotic manipulation represents a promising direction.

While existing research has focused on text-based task planning utilizing LLMs, there has recently been an emergence of general-purpose VLMs such as GPT-4V(ision) and GPT-4o. Integrating these vision systems into task planning opens up the possibility of developing task planners based on \textit{multimodal human instructions}. However, to the best of our knowledge, there has been only limited development of pipelines for multimodal task planners that combine off-the-shelf VLMs.

\begin{figure}[t]
  \centering
  \includegraphics[width=0.38\textwidth]{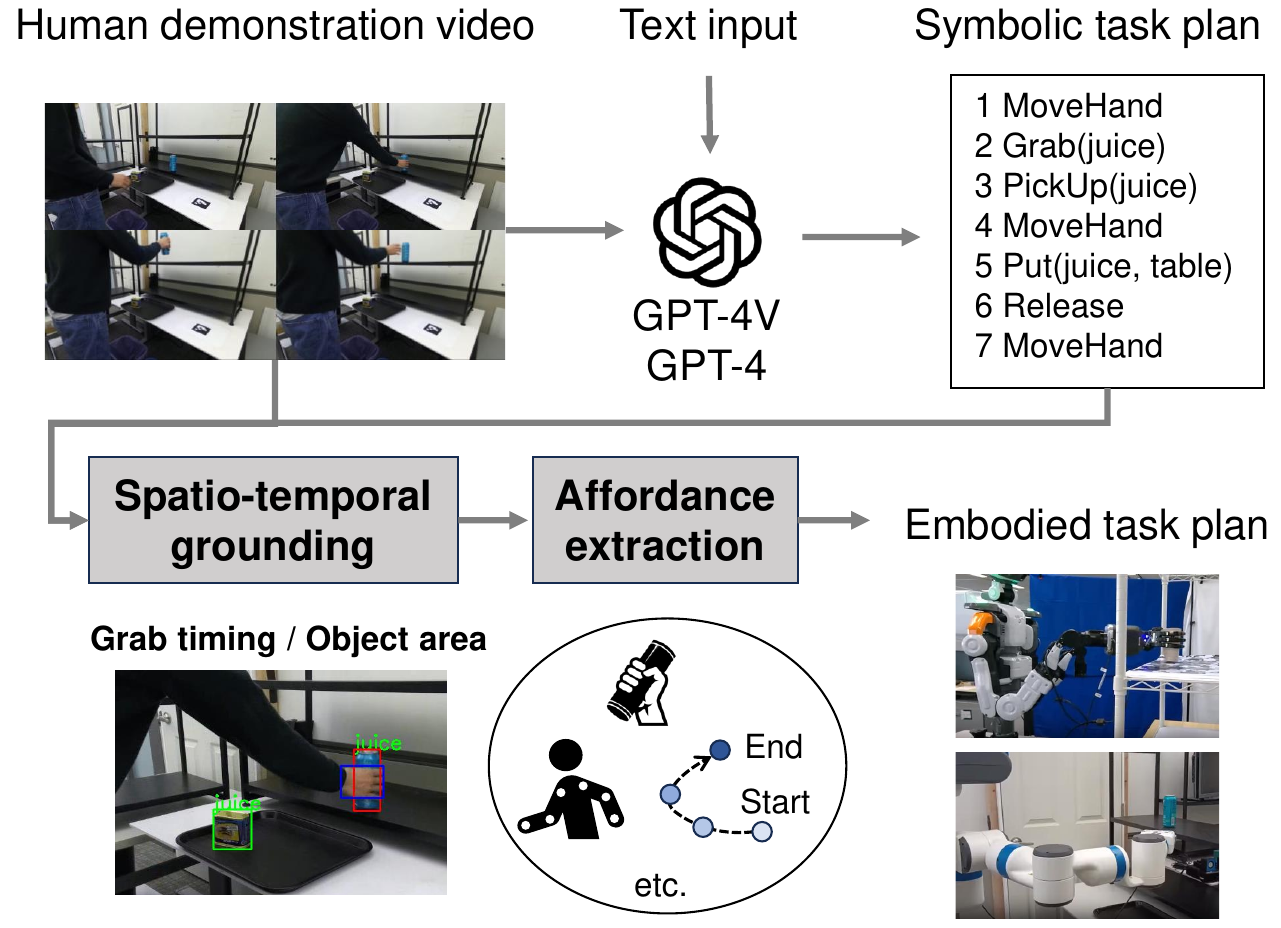}
  \caption{The proposed multimodal task planner highlights the ability to process video demonstrations and text input to generate task plans and extract key affordances for robot execution.}
  \label{fig:overview}
  \vspace{-7mm}
\end{figure}

This study proposes a multimodal task planner utilizing GPT-4V and GPT-4 (Fig.~\ref{fig:overview}), as examples of the recent VLM and LLM, respectively. This system accepts either human video demonstrations, text instructions, or both, and outputs symbolic task plans (i.e., a sequence of coherent task steps). When the visual data is available, the system then re-analyzes the videos in consideration of the task plan and identifies spatiotemporal correspondences between each task and the video. This process enables the extraction of various affordance information valuable for robot execution, such as the hand's approach direction when grasping objects, grasp types, collision-avoiding waypoints, and upper limb postures. Finally, the affordance information and the task plan are compiled into a hardware-independent executable file in the JSON format. 

We qualitatively evaluated the pipeline and confirmed the operability of the output task plan on several real robots. Additionally, we quantitatively tested the pipeline using a public cooking video dataset, which we manually labeled with tasks for robotic manipulation. Although the results were promising, we observed instances of hallucination in GPT-4V as a limitation of the model, highlighting the importance of incorporating human supervision within the pipeline.

This research makes three contributions: (1) Proposing a ready-to-use multimodal task planner that utilizes off-the-shelf VLM and LLM; (2) Proposing a methodology for aligning GPT-4V's recognition with affordance information for grounded robotic manipulation; (3) Making the code publicly accessible as a practical resource for the robotics research community.

\section{Related work}
\subsection{LLM/VLM-based task planning}
A methodology to operate robots from instructions has been a research topic before the emergence of LLMs~\cite{pramanick2020decomplex, venkatesh2021translating, yanaokura2022multimodal}. Recent studies aim to leverage the LLM/VLMs technologies~\cite{jiang2022vima, shridhar2023perceiver, brohan2023can, huang2022inner, ding2023task, singh2023progprompt,namasivayam2023learning, zhao2023differentiable,ding2022robot,zeng2022socratic,liang2023code,raman2022planning,xie2023translating}, and a large body of these studies aim to train an end-to-end custom model using specific datasets~\cite{brohan2023rt,brohan2022rt,li2023vision,ahn2022can,shah2023mutex,li2023mastering, jiang2022vima,shridhar2023perceiver,lynch2020language,brohan2023can,pan2023dataefficient,lin2023text2motion,zhao2023erra,liu2023instructionfollowing,namasivayam2023learning,zhao2023differentiable,mees2023grounding}. For example, Brohan et al.~\cite{brohan2023rt} proposed a transformer-based model that is trained based on both robotic trajectory data and
internet-scale vision-language tasks. However, such approaches often require a large amount of robotic data collected by experts and  necessitate data recollection and model retraining when transferring or extending these to other robotic settings. 

On the other hand, studies utilizing off-the-shelf LLMs focus on decomposing human instructions into high-level subgoals, while pre-trained skills achieve the subgoals~\cite{zhou2023generalizable, xu2023creative, sun2023prompt}. This approach is typically seen as a part of a planning framework, called task and motion planning (TAMP)\cite{garrett2021integrated}. 

This study is positioned as a part of the TAMP-based approach while extending the method to multimodal input by leveraging off-the-shelf GPT-4V and GPT-4.

\subsection{Grounding visual information for robotics}
The advanced language processing abilities of LLMs demonstrate the capability to interpret instructions and decompose them into robot action steps~\cite{ni2023grid, li2023interactive, parakh2023human,wake_chatgpt}. However, executing long task steps as planned is often challenging due to unforeseen and unpredicted environmental situations. Thus, one challenge in robotics is grounding task plans in environmental information. For example, there are approaches that focus on enabling LLMs to output the pre-conditions and post-conditions (e.g., states of objects and their interrelationships) of task steps to optimize their execution~\cite{zhou2023generalizable} and detect pre-condition errors for necessary revisions to the task plan~\cite{raman2023cape}. 
These strategies seek to achieve environment-grounded robot execution by integrating environmental information and adjusting the robot's actions at the task plan or controller level.

In this study, an open-vocabulary object detector~\cite{zhou2022detecting} is used to localize objects whose names are generated by GPT-4V. Additionally, by focusing on the interaction between the hand and the identified objects, the system can detect the timing and locations of grasping and releasing. This information aids vision systems in estimating affordances (e.g., grasp types, waypoints, and body postures) from human demonstration. The affordance information will be used when robots execute the tasks.
\subsection{Learning affordance}
The concept of affordance, as defined by Gibson~\cite{gibson2014ecological}, refers to the potential for action that objects or situations in an environment provide to an agent. In the field of robotics, the term `affordance' often refers to two key concepts: executable actions in an environment and the actionable space. For instance, Ahn et al.~\cite{ahn2022can} have proposed an approach that calculates the feasibility of robotic functions from visual information and compares it with planned tasks. Huang et al.~\cite{huang2023voxposer} proposed using LLMs/VLMs to extract the knowledge of movable areas. More recent approaches have proposed using off-the-shelf VLMs to identify language-aligned regions in given images~\cite{liu2024moka,huang2024manipvqa,chen2024spatialvlm,nasiriany2024pivot}.

These studies indeed define affordance as a form of Gibson's affordance; however, the term encompasses more, particularly regarding the interactions between the working environment and the objects being manipulated~\cite{ikeuchi2024applying}. For example, the notion of affordance can be extended to waypoints for collision avoidance~\cite{wake2022interactive}, grasp types~\cite{wake2023text}, and upper-limb postures~\cite{sasabuchi2020task}. This information is often not taught explicitly, thus vision systems need to extract it from human teaching demonstrations. In this study, we propose a pipeline to extract this information and provide a task plan endowed with that affordance information.

\section{Multimodal task planner}
\begin{figure*}[tbh]
  \vspace{2mm}
  \centering
  \includegraphics[width=0.85\textwidth]{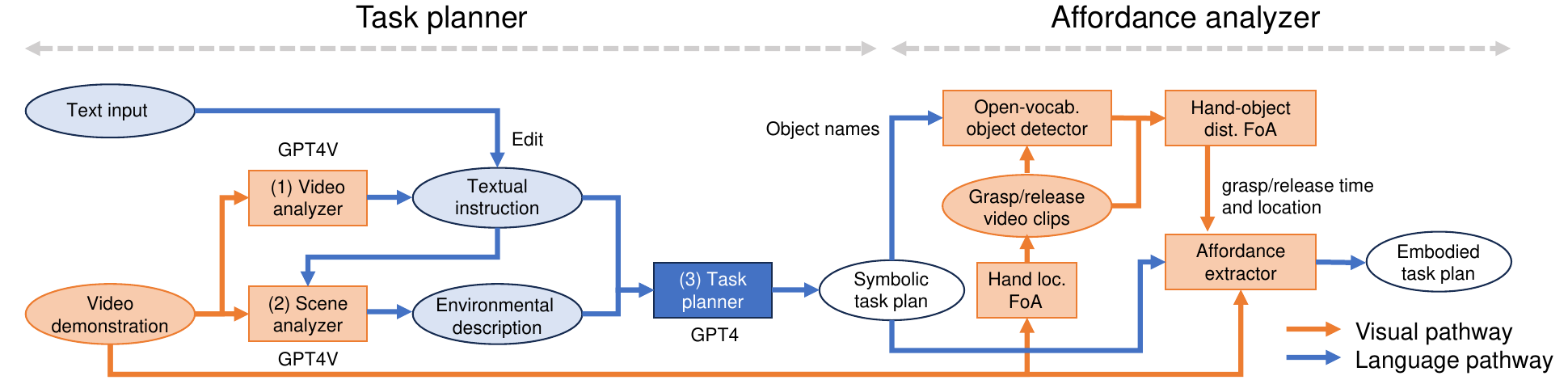}
  \vspace{-2mm}
  \caption{Proposed pipeline of the multimodal task planner. It consists of the symbolic task planner and the affordance analyzer. Blue components/lines are text-based information, and the red components are vision-related information. FoA denotes focus-of-attention.}
  \label{fig:pipeline}
  \vspace{-5mm}
\end{figure*}

The proposed system is composed of two pipelines connected in series (Fig.~\ref{fig:pipeline}). The first pipeline, the so-called symbolic task planner, takes teaching videos, text, or both as input, and then outputs a sequence of symbolic robot actions. Here, the text input includes feedback on the GPT-4V's recognition results for correction purposes. Providing users with the opportunity to give feedback on the recognition results enables robust operation. The second pipeline, the so-called affordance analyzer, analyzes the video to determine when and where the tasks occur, and then extracts the affordance information necessary for efficient task executions. Notably, GPT-4V is not used for the vision-analysis components within the affordance analyzer.

In this system, we assume that the robot operates in the same environment where humans demonstrate tasks. We also allow for slight object shifts, provided they can be corrected by a vision system during robot operation. Furthermore, for the experiments in this study, we use videos that capture the granularity of the grasp-manipulation-release operation, which starts with preparatory actions for grasping an object and ends with its release~\cite{wake2020learning}. Part of the prompt is shown in the figures, and the whole prompt is available at 
\href{https://microsoft.github.io/GPT4Vision-Robot-Manipulation-Prompts/}{https://microsoft.github.io/GPT4Vision-Robot-Manipulation-Prompts/}
\subsection{Symbolic task planner}

\begin{figure}[t]
  \centering
  \includegraphics[width=0.40\textwidth]{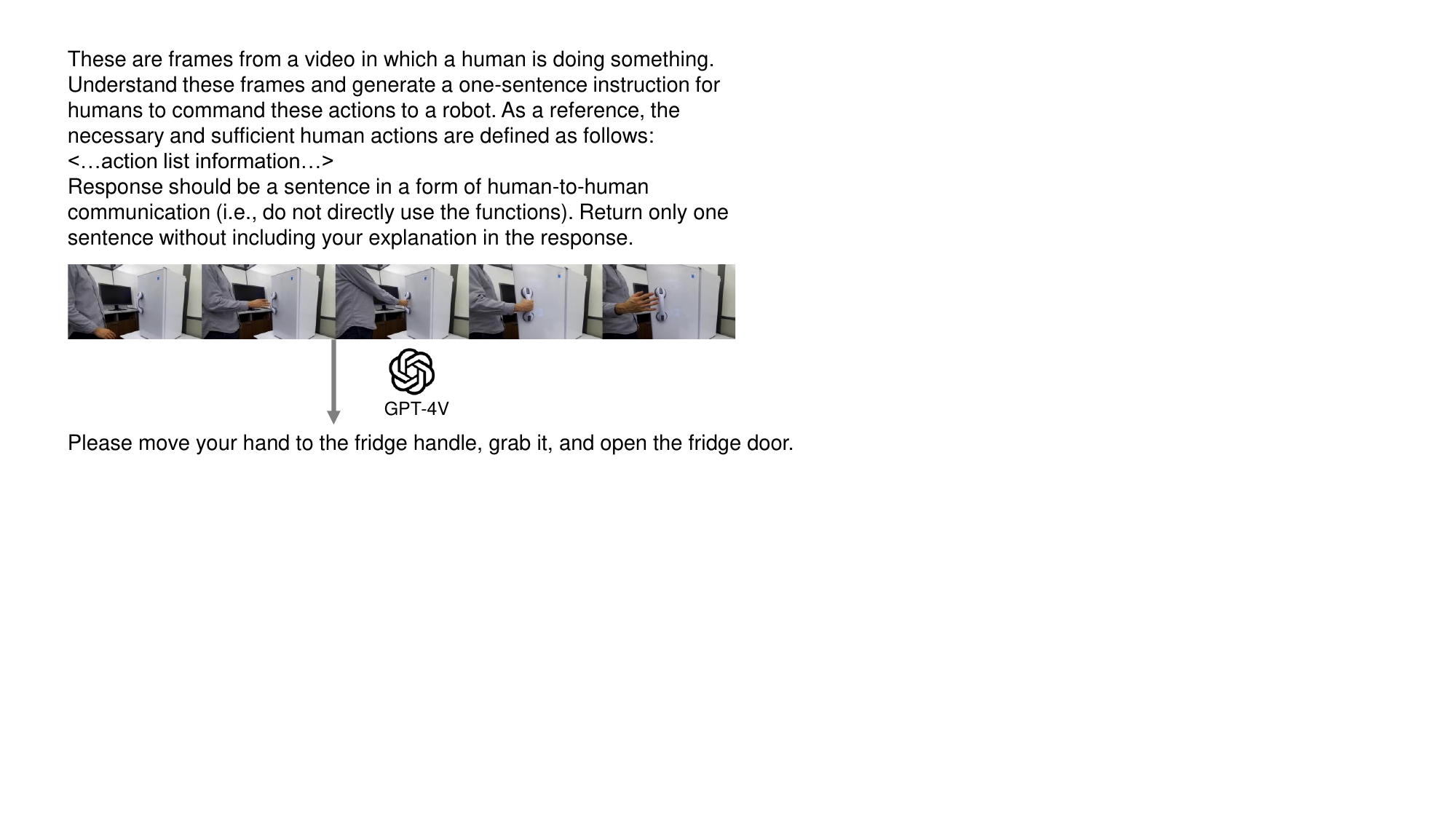}
  \vspace{-3mm}
  \caption{An example of the video analyzer's operation. The prompt and images fed into GPT-4V are shown at the top, and the corresponding output is shown at the bottom.}
  \label{fig:video-analyzer}
  \vspace{-7mm}
\end{figure}

The symbolic task planner is built from three components: 1) the video analyzer, 2) the scene analyzer, and 3) the task planner. Initially, given an RGB video, the video analyzer uses GPT-4V to recognize the actions performed by humans in the video and transcribes them into text instructions in a style used in human-to-human communication (e.g., `Please throw away this empty can'). In the video analysis, considering the model's token limit and latency, frames are extracted at regular intervals instead of from every frame, and then fed into GPT-4V. We used five frames as a practical number for processing. The output text is then checked and edited by the user. Figure~\ref{fig:video-analyzer} shows an example of the video analyzer's operation, suggesting that GPT-4V can successfully understand the human action from the frames. 

\begin{figure}[t]
  \centering
  \includegraphics[width=0.45\textwidth]{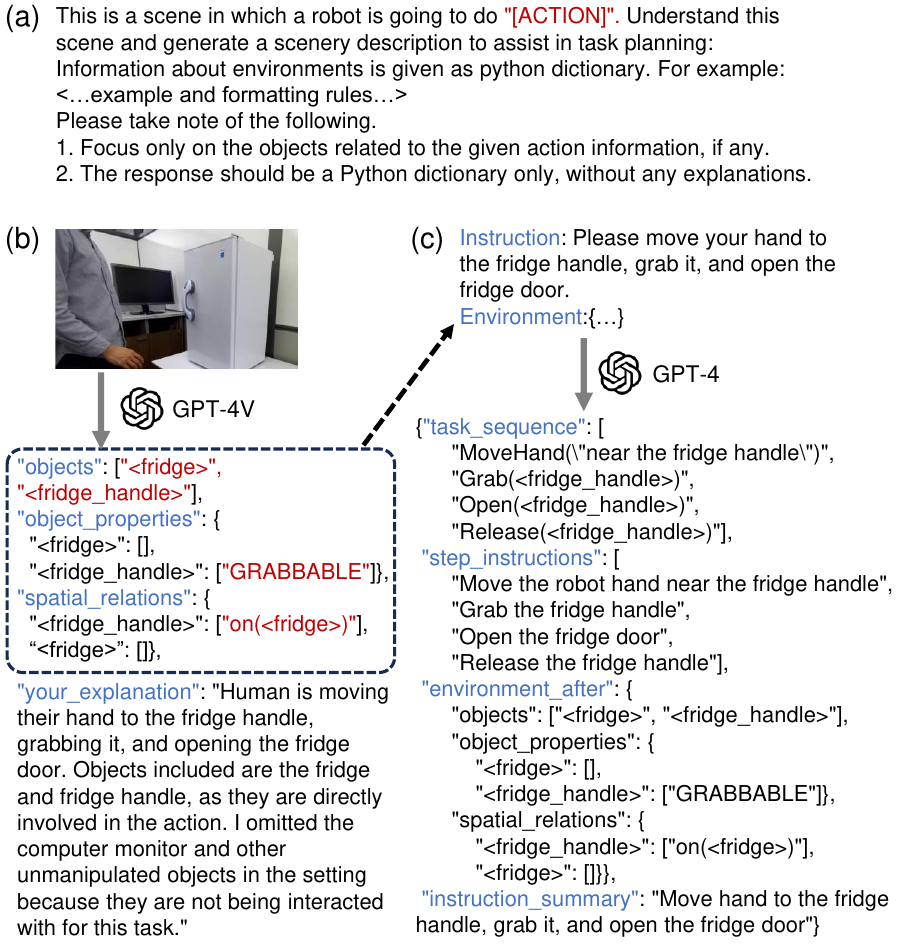}
  \vspace{-3mm}
  \caption{(a) The prompt for the scene analyzer. The input to GPT-4V is the first frame of the video and the textual instruction, which is replaced with "[ACTION]" in the prompt. Refer to Fig.~\ref{fig:video-analyzer} for an example of textual instruction. (b) The examples of its output. (c) Output of the task planner.}
  \label{fig:taskplanner}
  \vspace{-3mm}
\end{figure}

Next, the scene analyzer encodes the text instructions and the first frame of the video into the environmental description of the working area. This description includes a list of object names recognized by GPT-4V, the graspable properties of objects, and the spatial relationships between objects. Figure~\ref{fig:taskplanner} (a, b) shows the prompt and an example of the output from the scene analyzer. In this example, the output includes a fridge handle when opening a fridge, while ignoring a computer display in the background. These results suggest that the scene analyzer can effectively encode scene information in light of human actions. We prompted GPT-4V to explain the results of the object selection process and the reasons behind those choices, which led to robust outputs.

Finally, the task planner outputs task sequences from text instructions and environmental descriptions. Specifically, we designed a pre-prompt to make GPT-4 decompose the given instruction into a sequence of robot tasks~\cite{wake_chatgpt}. The set of robot tasks was defined based on the change in the motion constraints on an object being manipulated~\cite{ikeuchi2024semantic}, following the Kuhn-Tucker theory~\cite{kuhn1956related}. This definition allows us to theoretically establish a necessary and sufficient set of robot actions for object manipulation. Table~\ref{tab:robot_task_list} shows the set of tasks and the explanations that we included in the pre-prompt. Note that the object names are given in an open-vocabulary format based on the understanding by GPT-4V, and that the objects are identified in the video by the affordance analyzer at a subsequent stage.
Additionally, since the fine-grained robot task planning is directed by pre-prompting, users do not need to prepare a dataset for training the model---extending the action set only requires modifying the pre-prompt.

To ensure transparency in GPT-4's understanding, the task planner is designed to output explanations for the tasks, estimated environments after the operation, and the summary of the operation, as well as a set of task plans. Additionally, the task planner is a stateful system that maintains a history of past conversations within the token limits of the GPT-4. Therefore, users can modify and confirm the output through linguistic feedback~\cite{wake_chatgpt}. Figure~\ref{fig:taskplanner} (c) shows an example of the output from the task planner, suggesting that a set of tailored prompts results in reasonable textual instruction, environmental description, and the symbolic task plan.

\begin{table}[ht]
\vspace{3mm}
\caption{Robot actions and their explanations}
\centering
\vspace{-3mm}
\resizebox{0.45\textwidth}{!}{
\begin{tabular}{|l|p{6cm}|}
\hline
\textbf{Action} & \textbf{Description} \\
\hline
Grab(arg1) & Take hold of arg1. Preconditions: Arg1 is within reachable distance and no object is currently held. Postconditions: Arg1 is being held. \\
\hline
MoveHand(arg1) & Move the robot hand closer to arg1, where arg1 describes the hand's destination. \\
\hline
Release(arg1) & Release arg1. Preconditions: Arg1 is currently being held. Postconditions: Arg1 is no longer held. \\
\hline
PickUp(arg1) & Lift arg1. Preconditions: Arg1 is currently being held. Postconditions: Arg1 continues to be held. \\
\hline
Put(arg1, arg2) & Place arg1 onto arg2. Preconditions: Arg1 is currently being held. Postconditions: Arg1 continues to be held. \\
\hline
Rotate(arg1) & Open or close something by rotating arg1 along an axis. Preconditions: Arg1 is currently being held. Postconditions: Arg1 continues to be held.\\
\hline
Slide(arg1) & Open or close something by linearly moving arg1 along an axis. Preconditions: Arg1 is currently being held. Postconditions: Arg1 continues to be held.\\
\hline
MoveOnSurface(arg1) & Move arg1 across a surface. Preconditions: Arg1 is currently being held. Postconditions: Arg1 continues to be held.\\
\hline
\end{tabular}
}
\label{tab:robot_task_list}
\vspace{-8mm}
\end{table}

\subsection{Affordance analyzer}
The affordance analyzer re-analyzes the given videos using the knowledge from the symbolic task planner to acquire the affordance information necessary for the robot's effective execution. Specifically, it focuses on the relationship between hands and objects based on the task's nature and object names. It identifies the moments and locations of grasping and releasing in the video, aligning these with the task sequence. These moments serve as anchors for recognizing the affordances essential for each task. The effectiveness of focus-of-attention in detecting action grasping and releasing has been demonstrated in our prior study~\cite{wake2020verbal}.
\subsubsection{Attention to human hands to detect grasping and releasing}
Initially, the pipeline divides a series of videos into video clips at regular time intervals. The beginning and end frames of each video clip are then analyzed using a hand detector and an image classifier that determines whether an object is being grasped or not. The clips are classified into the following patterns:
\begin{itemize}
    \item Grasp video clip: clips where nothing is held in the first frame, but something is grasped in the last frame.
    \item Release video clip: clips where something is held in the first frame and nothing is held in the last frame.
    \item Other clips: clips that fall outside the definitions above.
\end{itemize}
This classification allows the analyzer to determine which video clips contain instances of grasping and releasing. For this purpose, we prepared a YOLO-based hand detector and hand recognizer~\cite{Ultralytics2023}.

\subsubsection{Attention to hand-object interaction to detect the spatiotemporal location of grasping and releasing}
\begin{figure}[t]
  \vspace{2mm}
  \centering
  \includegraphics[width=0.45\textwidth]{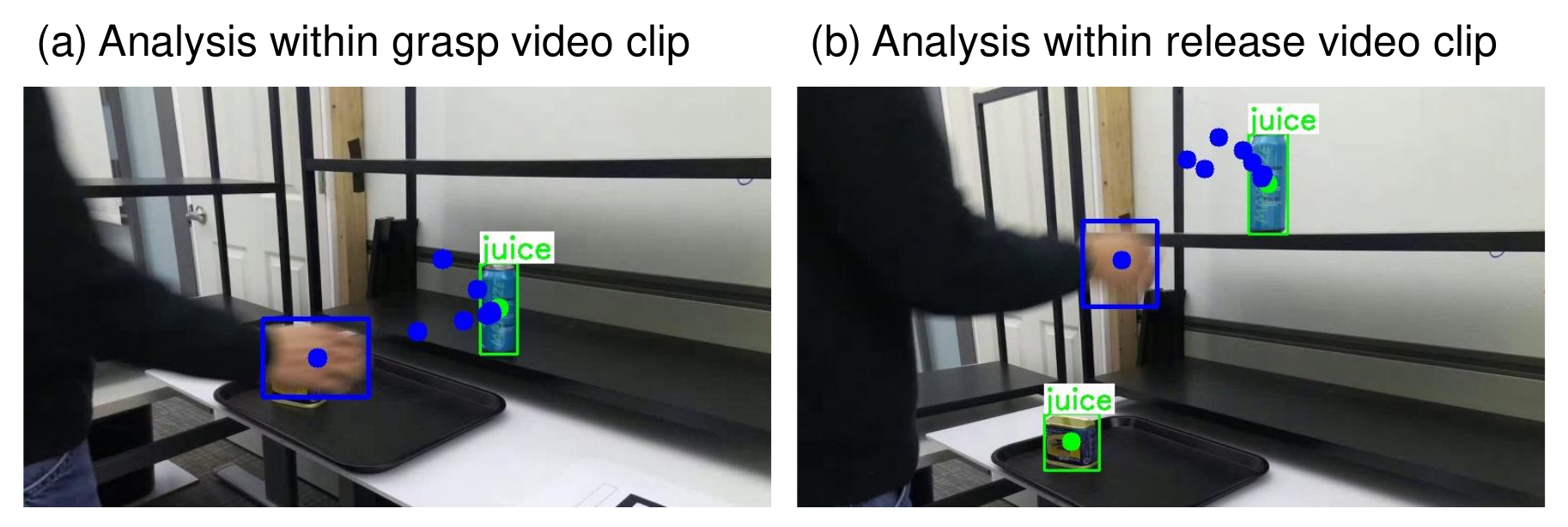}
  \vspace{-4mm}
  \caption{Detection of the objects by focusing on the relationship between the hand and the object. Green rectangles are the candidates for the object detected by the Detic model. The hand positions in the video clip are marked with blue points. Images are cropped for visualization purposes.}
  \label{fig:object_detection}
  \vspace{-3mm}
\end{figure}

\begin{figure}[t]
  \centering
  \includegraphics[width=0.45\textwidth]{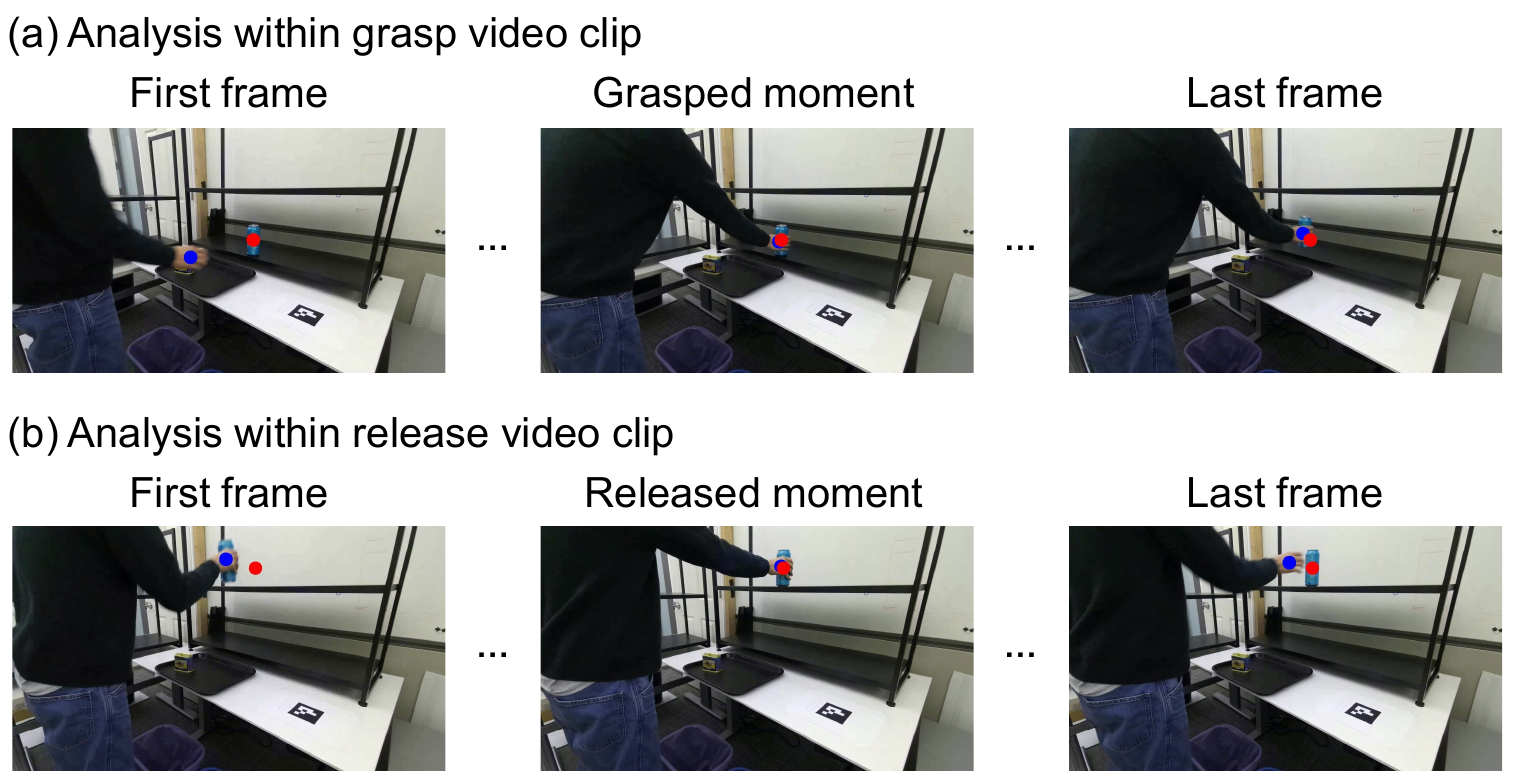}
  \vspace{-4mm}
  \caption{Detection of the timings and locations of grasping and releasing by focusing on the relationship between the hand and the object. The human hand moved a juice can from the bottom to the top of the shelf. The centers of the hand and the object are marked with blue and red points, respectively.}
  \label{fig:foa}
  \vspace{-5mm}
\end{figure}

The pipeline then focuses on the grasp video clip, analyzing the position and timing of the grasped object. We use Detic, an off-the-shelf, open-vocabulary object detector~\cite{zhou2022detecting}, to search for object candidates within the video, as identified in the symbolic task planner. When multiple object candidates are detected, the one closest to the hand in the video clip is deemed the grasped object. This is determined by comparing the distance between the bounding boxes of each candidate and the hand, as detected by the hand detector in every frame of the grasp video clip. Figure~\ref{fig:object_detection} illustrates the computation of object detection. The moment when the hand and the object are spatially closest during the grasp clip is identified as the moment of grasping. A similar calculation is applied to the release video clips to determine the timing of release. Figure~\ref{fig:foa} shows an example of the computation for moving a juice can from the bottom to the top of a shelf.

\subsubsection{Extracting affordance from aligned videos}
The moments of grasping and releasing serve as anchors to align the task sequence with the video. Following this alignment, the vision analyzer extracts affordance information of the object manipulation, including:

\begin{itemize}
    \item Grab task: 1) The hand's approach direction towards the object to avoid collisions with the environment. 2) The grasp type, based on the closure theory~\cite{yoshikawa1996}, also contains knowledge about how humans efficiently perform manipulations.
    \item MoveHand task: 1) The waypoints of the hand to avoid collisions with the environment.
    \item Release task: 1) The hand's withdrawal direction after releasing the object to avoid collisions with the environment.
    \item PickUP task: 1) The hand's departure direction to minimize unnecessary forces between the object and the plane.
    \item Put task: 1) The hand's approach direction towards the plane to avoid collisions with the environment.
    \item Rotate task: 1) The direction of the rotation axis. 2) The position of the rotation center. 3) The angle of rotation.
    \item Slide task: 1) The displacement of the sliding motion.
    \item MoveOnSurface task: 1) The axis that is vertical to the surface.
\end{itemize}
In addition to these affordances, the upper arm and forearm postures at the moments of grasping, releasing, and each waypoint are encoded as pairs of discretized directional vectors~\cite{wake2020learning}. These serve as constraints for computing Inverse Kinematics in multi-degrees-of-freedom arms~\cite{sasabuchi2020task}, ensuring that robot postures are predictable for users nearby. Notably, although these affordances generally provide viable information for robotic controllers, additional data such as force feedback may be required during robot execution~\cite{wake2020learning}.

\section{Experiments}
\subsection{Qualitative results}
\begin{figure}[t]
\centering
\vspace{2mm}
\includegraphics[width=0.48\textwidth]{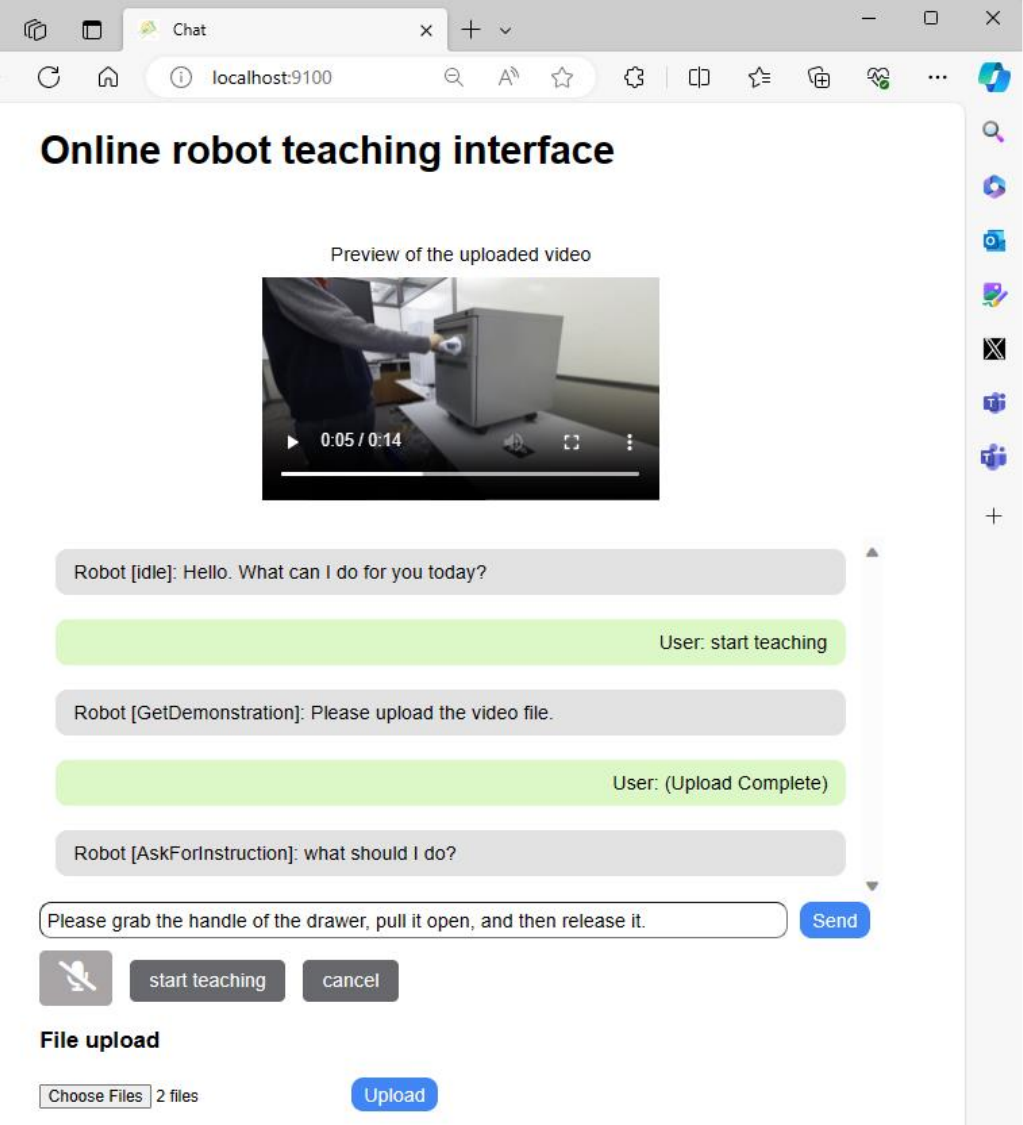}
\vspace{-7mm}
\caption{A web interface to operate the proposed pipeline. This interface allows users to upload pre-recorded teaching demonstrations, edit the results, and provide text feedback to GPT-4 and GPT-4V.}
\label{fig:webui}
\end{figure}

\begin{figure}[t]
\vspace{-2mm}
\centering
\includegraphics[width=0.45\textwidth]{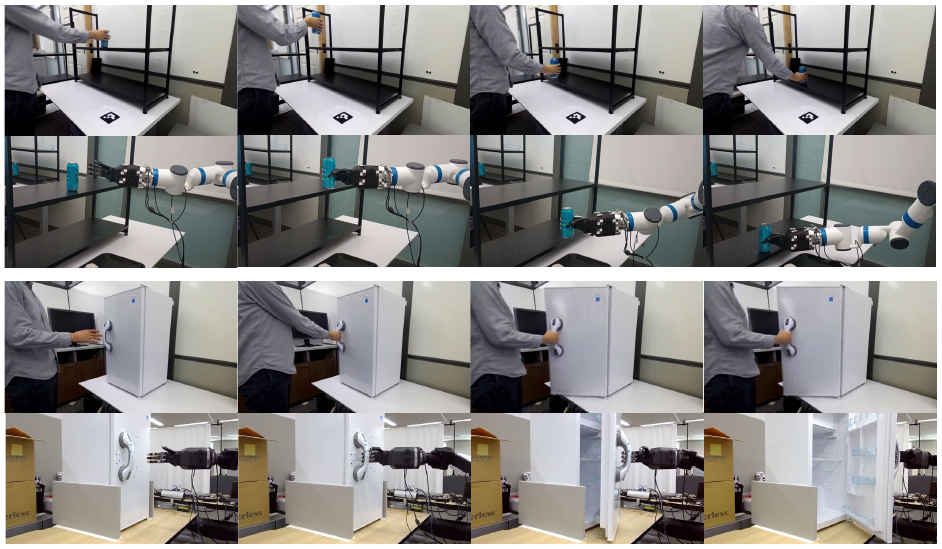}
\vspace{-2mm}
\caption{Examples of the robot execution from human demonstration. (Top) Relocating the juice between shelves. (Bottom) Opening a fridge. All results are available at \href{https://microsoft.github.io/GPT4Vision-Robot-Manipulation-Prompts/}{https://microsoft.github.io/GPT4Vision-Robot-Manipulation-Prompts/}}
\label{fig:robot}
\vspace{-6mm}
\end{figure}

We implemented the proposed pipeline as a single web interface (Fig.~\ref{fig:webui}). This interface allows users to upload pre-recorded teaching demonstrations, edit the results, and provide text feedback to GPT-4 and GPT-4V. 
The demonstration videos included a depth channel to aid in extracting spatial affordances, such as hand waypoints and approach directions.
We then tested whether the robot could be operated from videos obtained in various scenarios in a single attempt. Figure~\ref{fig:robot} shows several examples of its execution. Two robots were tested for the experiment: a Nextage robot (Kawada Robotics)\footnote{https://nextage.kawadarobot.co.jp/} with six degrees-of-freedom (DOF) in its arms, and a Fetch Mobile Manipulator (Fetch Robotics)\footnote{https://fetchrobotics.com/} with seven DOF in its arm. A four-fingered robot hand, the Shadow Dexterous Hand Lite (Shadow Robotics)\footnote{https://www.shadowrobot.com/dexterous-hand-series/}, was attached to the robots. 

Robot execution utilized our in-house control system~\cite{sasabuchi2023task}, equipped with skill libraries for grasping~\cite{saito2022task} and manipulation~\cite{takamatsu2022learning}. Specifically, the system used a first-person RGB-D camera mounted on each robot's head to locate objects via Detic. The system then selected a reference path, based on the demonstration, to guide the robot hand to the object. This path was chosen from pre-defined options for each robot hand to accommodate human-robot morphological differences. A reinforcement learning policy, specific to each grasp type, trained with randomized object positions and shapes, adjusted the hand posture to handle uncertainties~\cite{saito2022task}.

We confirmed that the robot operated not only in the demonstrated environment but also in an environment with a different texture, suggesting the system's robustness to environmental shifts. Additionally, we observed the successful reuse of `moving a juice can between shelves' scenario with multiple identical juice cans on the table, highlighting the reusability of this approach. 
All the experimental results are available at \href{https://microsoft.github.io/GPT4Vision-Robot-Manipulation-Prompts/}{https://microsoft.github.io/GPT4Vision-Robot-Manipulation-Prompts/}.

\subsection{Quantitative evaluation}
\begin{figure}[t]
  \vspace{2mm}
  \centering
  \includegraphics[width=0.45\textwidth]{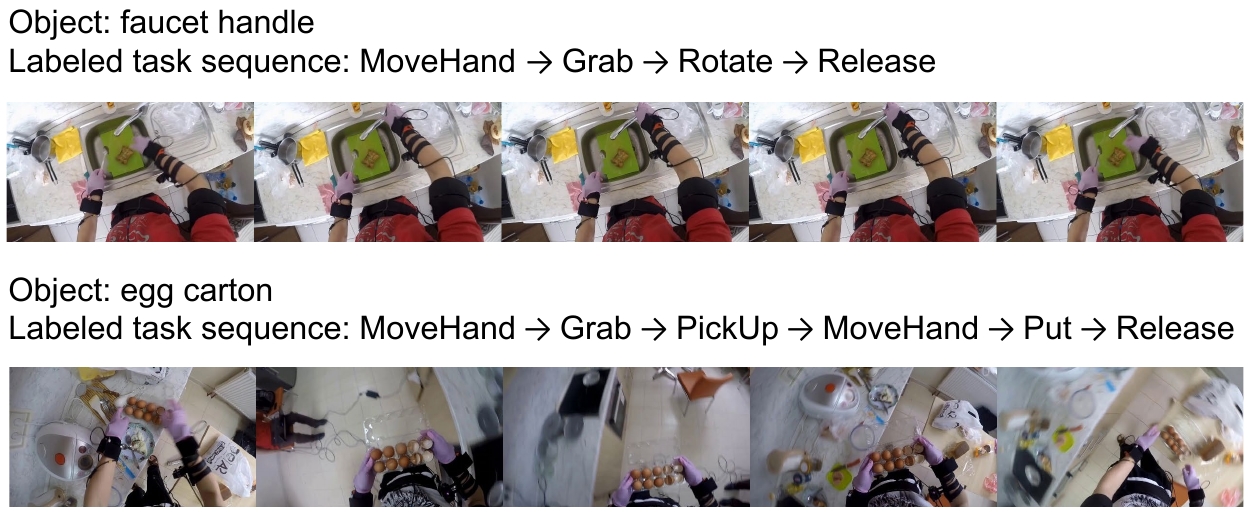}
  \vspace{-3mm}
  \caption{Example videos used for the quantitative tests and their labeled task sequences.}
  \label{fig:quantitative_test_overview}
  \vspace{-3mm}
\end{figure}

We evaluated our pipeline quantitatively using an existing dataset of first-person videos of human cooking activities~\cite{saudabayev2018human} (Fig.~\ref{fig:quantitative_test_overview}). We chose the cooking domain because it represents one of the most challenging domestic tasks in terms of image analyses due to the diverse objects cluttered in various environments. First-person videos were adopted to minimize the impact of visual occlusions. Existing datasets of human actions are labeled with coarse categories (e.g., preparing pasta) and simple motion labels often include adjacent manipulations (e.g., a motion labeled as `pick,' includes grasping and carrying~\cite{goyal2017something, damen2020epic}). To our knowledge, no existing datasets label tasks based on the changes in constraints of manipulated objects to suit the granularity of robot
actions~\cite{wake2020learning, ikeuchi2024semantic}. Therefore, we manually annotated a small subset of the cooking dataset with a series of task labels using a third-party annotation tool~\cite{video-annotator}. 

To this end, we identified grasp-manipulation-release operations that were composed of tasks in Table~\ref{tab:robot_task_list}. The annotators focused only on the right-hand manipulation, and 58 videos were analyzed. We then compared the output of the symbolic task planner with the annotated task sequences. This paper focuses on task recognition using off-the-shelf models without training. To our knowledge, there are no existing methods that output robot-task-level action sequences without involving model training. Therefore, we examined the performance of our proposed pipeline without comparing it to existing applications. 
\subsubsection{Performance of the video analyzer}
\begin{figure}[t]
  \centering
  \includegraphics[width=0.45\textwidth]{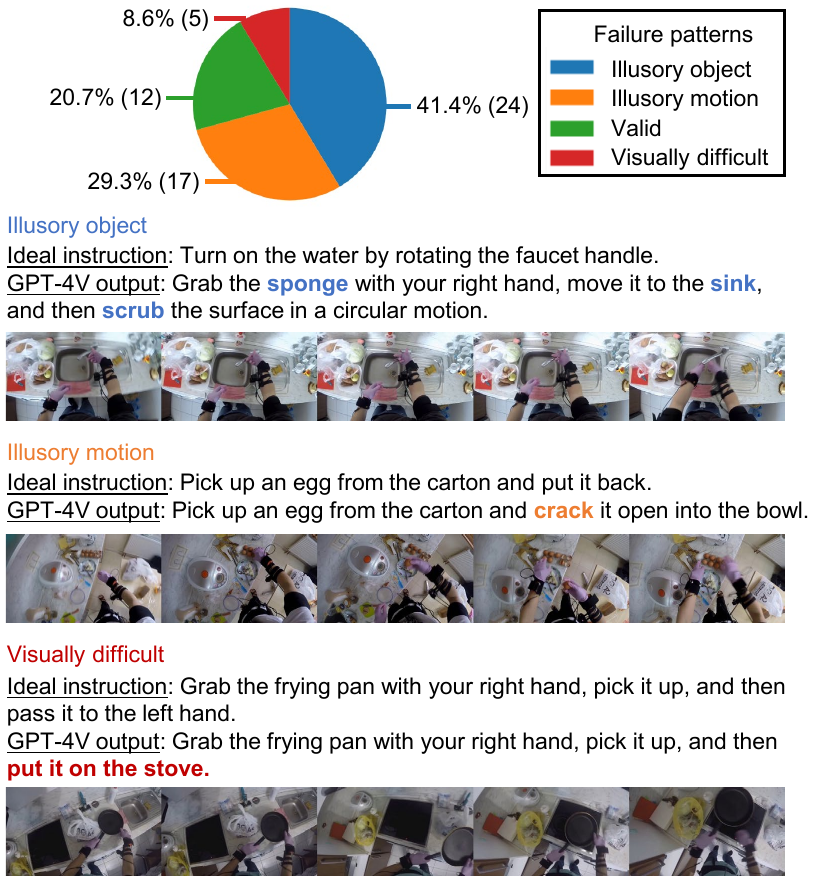}
  \vspace{-3mm}
  \caption{Performance of GPT-4V in recognizing cooking videos.}
  \label{fig:quantitative_test_result1}
  \vspace{-5mm}
\end{figure}

We manually checked the output of the video analyzer to understand the performance of GPT-4V (Fig.~\ref{fig:quantitative_test_result1}; Valid cases). The results showed that only a limited portion of the videos (20.7\%) were correctly transcribed in terms of the selection of the manipulated object's name and the action. We then analyzed the failure cases and determined three failure patterns (Fig.~\ref{fig:quantitative_test_result1}).
\begin{itemize}
    \item Illusory object: GPT-4V selected an incorrect object in or likely in the scene.
    \item Illusory motion: The manipulated object was correct but GPT-4V described incorrect tasks that can be associated with the object.
    \item Visually difficult: The output was incorrect but it was presumably due to limited spatial or temporal resolution, or perspective differences.
\end{itemize}

These failures in the image descriptions are not new and are often referred to as `hallucination' of VLMs~\cite{li2023evaluating}, which highlights the importance of humans' supervision (Fig.~\ref{fig:pipeline}) with the GPT-4V to date.

\subsubsection{Performance of the symbolic task planner}
We evaluated the quality of the output plan against the true action labels. Assuming that correct instructions are provided under human supervision, we manually collected outputs from the video analyzer and tested 58 videos. The alignment between the output task sequence and the correct task sequence was quantified using the normalized Levenshtein distance, which ranges from 0 to 1, with 1 indicating a perfect match. We compared three pipelines to assess the impact of the scene analyzer and human feedback on task planning: one with both the scene analyzer and correct instructions, one with only correct instructions, and one with neither (Table~\ref{tab:task_planning}). The comparison between pipelines with and without correction (0.76 vs 0.87) highlights the effectiveness of involving humans in the loop. Incorporating the scene analyzer improved performance, suggesting that scene information about the manipulated object (Fig.~\ref{fig:taskplanner}) can guide text-based task planning through general knowledge of an object's manipulability and the relationships between objects.
\vspace{-1mm}
\begin{table}[h!]
    \caption{Task-planning performance given a correct instruction}
    \vspace{-3mm}
    \centering
    \resizebox{0.40\textwidth}{!}{
        \label{tab:task_planning}
            \begin{tabular}{lcc}
            \toprule
            \textbf{Pipeline} & \textbf{Mean} & \textbf{Standard deviation} \\
            \midrule
            Task planner & 0.76 & 0.16 \\
            Task planner + FB & 0.87 & 0.12 \\
            Task planner + SA + FB& \textbf{0.90} & 0.11 \\
            \bottomrule
            \end{tabular}
        }
        \footnotesize
        \begin{tabular}[t]{p{0.45\textwidth}}
        \centering
          SA: Scene analyzer, FB: Human corrective feedback. 
        \end{tabular}
        \vspace{-3mm}
\end{table}

\subsubsection{End-to-end robot operation}
We evaluated the end-to-end success rates for the operations of moving a juice can between shelves and opening a drawer, using 20 human demonstrations for each. 
A SEED-noid robot (THK)\footnote{https://www.thk.com/jp/en/} with a one-DOF gripper was used for this experiment.
Table~\ref{tab:end_to_end} shows cases of successful operations without errors or collisions, as well as the cases where the symbolic task plans (Fig.~\ref{fig:pipeline}) were valid in terms of task sequence and object names.
An ablation study showed that the scene analyzer had minimal impact, likely due to the simplicity of the environment (Fig.~\ref{fig:robot}). We observed cases where GPT-4V misinterpreted instructions, such as confusing a drawer for a safe, resulting in failures (2 out of 20) in the columns labeled `w/o' and `w/ SA,' highlighting the importance of human feedback (column `w/ SA \& FB').
Detection failures in real robots stemmed from flawed affordance information, caused by inaccuracies in hand detection and image classification affecting grasp and release video clip identification. Notably, robot execution was tested solely on valid task plans for safety.
\vspace{-1mm}
\begin{table}[h!]
  \caption{End-to-end robot execution experiment}
  \vspace{-3mm}
  \centering
  \resizebox{0.45\textwidth}{!}{
      \label{tab:end_to_end}
      \begin{tabular}{lcccc}
      \toprule
      \multirow{3}{*}{\textbf{Operation}} & \multicolumn{3}{c}{\textbf{Valid Symbolic Task Plans}} & \multirow{3}{*}{\begin{tabular}[c]{@{}c@{}}\textbf{Successful} \\ \textbf{Executions}\textsuperscript{*}\end{tabular}} \\
      \cmidrule{2-4}
       & w/o SA & w/ SA & w/ SA \& FB & \\
      \midrule
      Juice Relocation & 20/20 & 20/20 & - & 19/20 \\
      Drawer Opening & 18/20 & 18/20 & 20/20 & 17/18 \\
      \bottomrule
      \end{tabular}
  }
  \footnotesize
  \begin{tabular}[t]{p{0.45\textwidth}}
  \centering
    SA: Scene analyzer, FB: Human corrective feedback. \\
    \textsuperscript{*}Robots were tested for valid symbolic task plans with SA.
  \end{tabular}
  \vspace{-3mm}
\end{table}

\section{Limitation and discussion}\label{discussion}

In this paper, we introduced a multimodal robot task planning pipeline utilizing GPT-4V. This pipeline interprets human actions in videos, integrates human textual feedback, and encodes relevant environmental information. High-level symbolic task plans are then generated by GPT-4. Following this planning phase, an open-vocabulary object detector spatially localizes the objects. A vision system identifies the moments and locations for grasping and releasing based on the interaction between hands and objects. This video reanalysis allows the system to extract affordance information that is useful for robot execution.

The real-robot experiments have demonstrated the effectiveness of this pipeline in various scenarios with the end-to-end performance reaching 85-95\% (Table~\ref{tab:end_to_end}), which is comparable to model-based approaches (e.g.,~\cite{brohan2023rt}). However, the quantitative evaluation revealed limited performance in video understanding by GPT-4V. Notably, the limited performance of GPT-4V can be attributed to two factors: the operations tested in kitchens cluttered with objects, and self-motion causing shifts and blurring of the images, which made recognition challenging for the model. Furthermore, the current approach of frame selection (i.e., five frames at regular intervals) can affect the model's ability to infer critical actions in the demonstration. These facts highlight the importance of incorporating human supervision and corrections into the loop or improving the model through prompt engineering.

Alongside the performance issues in video understanding, the system comes with several limitations:
\begin{itemize}
\item \textbf{Extension to long steps}: Given that the video grounding was computed only for the grasping and releasing moments, the affordance information extracted from the demonstration is inherently limited. Future work involves extracting additional affordance information, which entails the challenge of precisely localizing each task's action within the video.
\item \textbf{Higher-order pre- and post-conditions}: The pipeline primarily considered object relationships to determine pre- and post-task conditions. However, the criteria for task completion may extend beyond object relationships. For instance, a MoveOnSurface task for cleaning should involve the removal of dirt from the surface. Developing methodologies for GPT-4V/GPT-4 to consider these complex objectives requires further exploration.
\item \textbf{Optimization of prompts}: The success of task planning with VLMs/LLMs is heavily influenced by the design of prompts. Accumulating practical expertise in prompt engineering is crucial for the research community to enhance the effectiveness of these models.
\end{itemize}

\section{Conclusion}
We introduced a novel multimodal robot task planning pipeline utilizing GPT-4V, effectively converting human actions from videos into robot-executable programs. As the proposed task planner is based on off-the-shelf models, it offers flexibility in application across a wide range of robotic hardware and scenarios. We hope that this study will serve as a practical resource for the robotics research community and inspire further developments in this field.

\section*{Acknowledgment}
We thank Dr. Sakiko Yamamoto, Dr. Etsuko Saito (Ochanomizu University), and Dr. Midori Otake (Tokyo Gakugei University) for their help in annotating the cooking dataset. This study was conceptualized, conducted, and written by the authors, and OpenAI's GPT-4 was used for proofreading.

\bibliographystyle{ieeetr}
\bibliography{bib}
\end{document}